\documentclass[sigconf]{acmart}

\AtBeginDocument{%
  \providecommand\BibTeX{{%
    \normalfont B\kern-0.5em{\scshape i\kern-0.25em b}\kern-0.8em\TeX}}}

\copyrightyear{2024}
\acmYear{2024}
\setcopyright{acmlicensed}
\acmConference[WSDM '24]{Proceedings of the 17th ACM International Conference on Web Search and Data Mining}{March 4--8, 2024}{Merida, Mexico}
\acmBooktitle{Proceedings of the 17th ACM International Conference on Web Search and Data Mining (WSDM '24), March 4--8, 2024, Merida, Mexico}
\acmPrice{15.00}
\acmDOI{10.1145/3616855.3635799}
\acmISBN{979-8-4007-0371-3/24/03}
\usepackage{graphicx}
\usepackage{amsmath}
\usepackage{amssymb} 
\usepackage{float}
\usepackage{subfigure}
\usepackage{makecell}
\usepackage{multirow}
\usepackage{color}
\usepackage{pifont}
\usepackage{algorithm}
\usepackage{algorithmicx}
\usepackage{algpseudocode}
\usepackage{microtype}
\usepackage{CJKutf8}
\usepackage{lineno}
\usepackage{soul}
\usepackage{capt-of}

\begin{document}

\title{Exploiting Duality in Open Information Extraction\\ with Predicate Prompt}

\author{Zhen Chen}
\email{zhenchen21@m.fudan.edu.cn}
\affiliation{%
  \institution{Fudan University, \\ Shanghai Key Laboratory of Data Science}
    \city{Shanghai}
  \country{China}
}
\author{Jingping Liu}
\authornote{Corresponding Author.}
\email{jingpingliu@ecust.edu.cn}
\affiliation{%
  \institution{ East China University of Science and Technology}
    \city{Shanghai}
  \country{China}
}
\author{Deqing Yang$^*$}
\email{yangdeqing@fudan.edu.cn}
\affiliation{%
  \institution{Fudan University,\\ Shanghai Key Laboratory of Data Science}
    \city{Shanghai}
  \country{China}
}
\author{Yanghua Xiao }
\email{shawyh@fudan.edu.cn}
\affiliation{%
  \institution{Fudan University,\\ Shanghai Key Laboratory of Data Science}
    \city{Shanghai}
  \country{China}
}
\author{Huimin Xu}
\email{xuhuimin04@meituan.com}
\affiliation{%
  \institution{Meituan}
    \city{Shanghai}
  \country{China}
}
\author{Zongyu Wang}
\email{wangzongyu02@meituan.com}
\affiliation{%
  \institution{Meituan}
    \city{Shanghai}
  \country{China}
}
\author{Rui Xie}
\email{ rui.xie@meituan.com}
\affiliation{%
  \institution{Meituan}
    \city{Shanghai}
  \country{China}
}
\author{Yunsen Xian}
\email{ xianyunsen@meituan.com}
\affiliation{%
  \institution{Meituan}
    \city{Shanghai}
  \country{China}
}
\renewcommand{\shortauthors}{Zhen Chen et al.}

\begin{abstract}
Open information extraction (OpenIE) aims to extract the schema-free triplets in the form of (\emph{subject}, \emph{predicate}, \emph{object}) from a given sentence. 
Compared with general information extraction (IE), OpenIE poses more challenges for the IE models, {especially when multiple complicated triplets exist in a sentence. To extract these complicated triplets more effectively, in this paper we propose a novel generative OpenIE model, namely \emph{DualOIE}, which achieves a dual task at the same time as extracting some triplets from the sentence, i.e., converting the triplets into the sentence.}
Such dual task encourages the model to correctly recognize the structure of the given sentence and thus is helpful to extract all potential triplets from the sentence. 
Specifically, DualOIE extracts the triplets in two steps: 1) first extracting a sequence of all potential predicates, 2) then using the predicate sequence as a prompt to induce the generation of triplets. Our experiments on two benchmarks and our dataset constructed from Meituan demonstrate that DualOIE achieves the best performance among the state-of-the-art baselines. Furthermore, the online A/B test on Meituan platform shows that 0.93\% improvement of QV-CTR and 0.56\% improvement of UV-CTR have been obtained when the triplets extracted by DualOIE were leveraged in Meituan's search system.
\end{abstract}


\maketitle

\section{Introduction}
Open information extraction (OpenIE) plays an important role in a variety of downstream tasks, such as knowledge base construction \cite{dong2014knowledge}, question answering \cite{yan2018assertion} and summarization \cite{cao2018faithful}. 
OpenIE aims to extract schema-free triplets in the form of (\emph{subject}, \emph{predicate}, \emph{object}) from unstructured natural language, where subjects and objects are both called arguments. 

In recent years, most OpenIE systems were devised based on deep neural networks (DNNs), which could be divided into two main categories: tagging-based and generative methods \cite{ijcai2022p793}. 
The tagging-based solutions \cite{stanovsky-etal-2018-supervised} model OpenIE as a sequence labeling problem, where each token in the input is tagged as a subject, predicate or object. 
The generative solutions model OpenIE as a Seq2Seq problem. For example, IMoJIE \cite{kolluru-etal-2020-imojie} adopts an iterative generation mechanism to alleviate redundancy in extractions.Gen2OIE \cite{kolluru2022alignment} leverages two mT5 models \cite{xue2020mt5} to extract triplets by reconstructing multiple inputs. 
However, the previous models' inadequate understanding of diverse relations between arguments and the sentence's intricate structure hinder them to extract the complicated triplets effectively, which include the following three categories \cite{yu2021maximal}. 1) \emph{Overlapping triplets} refer to those triplets sharing the same element, as the first two triplets in Table \ref{tb:1} sharing the same subject ``Shea''; 2) \emph{Discontinuous triplet} has the element composed by two separate spans, as the second triplet's predicate in Table \ref{tb:1} that composed of ``was born'' and ``in'' from the input; 3) Two triplets are regarded as \emph{nested triplet} , if an element in one triplet contains other element or shares some words with the other triplet's element. For example, the predicates of the first two triplets in Table \ref{tb:1} share the same word ``was born''.
{
In addition, the complicated triplets also include \emph{implicit triplets} that are overlooked by current OpenIE research. 
The predicate in an implicit triplet is not explicitly mentioned in the sentence. As shown in Table \ref{tb:1}, the predicate ``is in'' of the third triplet is absent from the input. 
Implicit triplets often emerge in incomplete sentences of oral expression, such as user comments in real-world scenarios. Thus, the inefficiency in extracting implicit triplets would degrade the downstream task's performance severely.
}

\begin{table}[t]
\centering
\caption{A toy example of OpenIE where three triplets can be extracted from the input sentence.}
\scalebox{0.85}{
\begin{tabular}{llcc}
\toprule
\textbf{Sentence}: 
Shea was born on September 5, 1900 in San Francisco, California. \\ \midrule
\textbf{Triplets}:       \\
1. (Shea, was born on, [September 5, 1900])       \\ 
2. (Shea, was born in, [San Francisco, California])      \\
3. (San Francisco, is in, California)\\ \bottomrule
\end{tabular}}
\label{tb:1}
\end{table}

To address the challenges in OpenIE posed by these complicated triplets, in this paper we propose a novel generative OpenIE model \emph{DualOIE}, which achieves enhanced OpenIE mainly through learning an auxiliary dual task of converting the triplets into a sentence.
{Since tagging model could not introduce new words, resulting in inefficiency in handling implicit triplets, our DualOIE adopts the paradigm of generation instead of tagging.}
Trained under the constraint of the dual task, our model is encouraged to capture the diverse relations between the arguments and recognize the overlapping elements in the overlapping triplet by figuring out how these elements fit into the context of the input sentence.
In addition, the dual task also encourages the model to be more aware of word order and sentence structure, thus helping the model resolve the discontinuous triplet and nested triplet in the input sentence.
{
For implicit triplets, transforming them into sentence enables the model to comprehend and learn those informal expressions, leading to more accurate identification of implicit triplets in the extraction task.
}

In summary, our DualOIE is built with a dual framework with two task directions: the \textbf{S (sentence) to T (triplet)} direction corresponds to the primary extraction task, while the \textbf{T (triplet) to S (sentence)} direction corresponds to the dual task. The two directions share the same encoder, enabling the model to jointly learn from the two tasks.
Specifically, in the S to T direction, given the sentence's complicated structure, directly using a Seq2Seq model to generate triplets from the input may result in incorrect output, such as repetition or omission.  
Therefore, we split the primary extraction process (S to T) into two steps:
1) first extracting a sequence of potential predicates from the input; 2) then using the predicate sequence as a prompt and concatenating it with the input to generate all triplets. 
Note that a subject of a triplet might be the object of another triplet, causing ambiguity in the second step, thus we first extract predicates as the prompt rather than extracting subjects or objects as the prompt.

{
One important reason of previous OpenIE models' inefficiency in extracting implicit triplets is that, they are trained with the datasets only having less implicit triplets.
}
According to our statistics, {
there are only 3\% and 11\% implicit triplets in the OpenIE benchmark CaRB and SAOKE, respectively.}
{Besides, the two datasets only have 
less than 30 types of implicit predicates, which also limits the performance of the models trained with them.
The sparsity of implicit triplets in these two benchmarks is due to that, they were collected from Wikipedia and Baidu Baike, where the expressions in sentences are more formalized and complete than oral expressions in various Web platforms such as Meituan, which is a famous Chinese review and search platform for daily life.
} 
{In order to train and evaluate OpenIE models towards implicit triplet extraction, we constructed a dataset \emph{MTOIE} (MeituanOpenIE) from the real-world scenario of Meituan.}
{
MTOIE was collected from the massive user reviews on Meituan, annotated following the philosophy of previous OpenIE benchmarks and crowdsourced by experienced workers. 
}
To sum up, MTOIE has 54,060 sentences and 87,971 triplets, in which there are about 33\% implicit triplets and more than 2,000 types of implicit predicates.

This paper's contributions are summarized
as follows:

\noindent\textbf{1.} 
We propose a novel OpenIE model based on a joint dual framework with two task directions, 
{
which enables the model to learn under the mutual constraints of both tasks 
}
and further use predicate prompt to alleviate the omission and repetition in the \textbf{S to T} direction. To the best of our knowledge, this is the first work to introduce duality in OpenIE.

\noindent\textbf{2.} 
{
To evaluate OpenIE models on complicated triplet extraction, we also construct a high-quality dataset MTOIE that contains more than 29,000 implicit triplets and more than 2,000 types of implicit predicates, which exceeds current benchmarks by a significant margin in terms of both quantity and variety.
}

\noindent\textbf{3.} 
We conduct extensive experiments on two benchmark datasets and our MTOIE, to verify our model's advantage over all baselines, including ChatGPT. Furthermore, the online A/B test on the Meituan platform shows that the query view click-through rate (QV-CTR) and unique visitor click-through rate (UV-CTR) have increased by 0.93\% and 0.56\% respectively when our model's extraction results were deployed into the search system.

{\emph{\small The source code of DualOIE will be soon available at \textcolor{blue}{\url{https://github.com/ccczhen/DualOIE}}.} }

\begin{figure*}[ht]
\centerline{\includegraphics[width =0.85\linewidth]{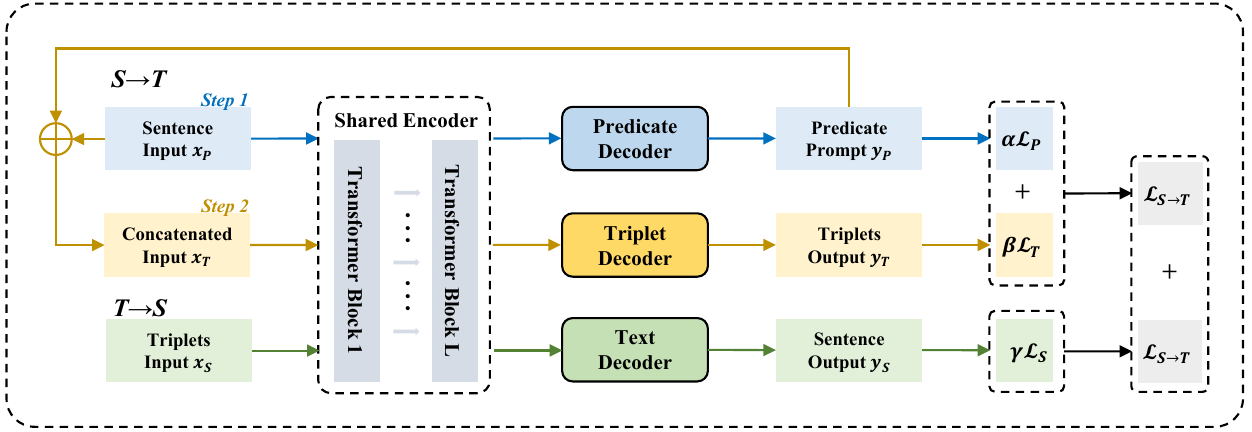}}
\caption{The overall framework of our proposed DualOIE, including the structure of achieving two tasks of opposite directions, $\textbf{S} \rightarrow \textbf{T}$ and $\textbf{T}\rightarrow \textbf{S}$.}\label{fig:framework}
\label{fig:structure}
\end{figure*}
\section{Overview}

In this section, we first formalize the problem of OpenIE task addressed in this paper, and then present the framework of our proposed DualOIE.
\subsection{Problem Definition}
Given an  input sentence $x=[w_1,...,w_n]$, where $w_i$ (1$\leq$$i$$\leq$$n$) is the $i$-th word (token), our task's goal is to extract a collection of triplets $Y$=$\{(s_i,p_i,o_i)\}_{i=1}^{|Y|}$ where $s_i,p_i,o_i$ denote the subject term, predicate term, object term of the $i$-th triplet, respectively. Generally, $s$ and $o$ might be a single word or a span extracted from $x$, and $p$ might even be absent in $x$. For example, given ``\emph{Joe Biden (November 20, 1942- ) is the U.S. president.}", a good model needs to output two triplets, i.e., (\emph{Joe Biden}, \emph{is}, \emph{the U.S. president.}) and (\emph{Joe Biden}, \emph{was born on}, \emph{[November 20, 1942]}). Note that the predicate of the second triplet is absent in the input sentence.

\subsection{Framework}
Given an input sentence $x$, the primary objective of our OpenIE model is to maximize the following probability
\begin{equation}
    p(Y|x),
    \label{eq:s2t}
\end{equation}
where $Y$ is the collection of fact triplets.
It is in fact the task of \textbf{S to T} direction mentioned before, i.e., extracting triplets from the input sentence. This direction contains two steps: 1) a prompt $z=[p_1,...,p_{|Y|}]$ is first extracted from $x$ as a prompt, where $p_i$ (1$\leq$$i$$\leq$$|Y|$) is the predicate of the $i$-th triplet in $Y$; 2) then $x$ is concatenated with $z$ as $[z;x]$ to generate $Y$. 

As the dual task of \textbf{S to T}, the \textbf{T to S} direction aims to convert the triplets into a sentence, of which the objective is to maximize
\begin{equation}
    p(x|Y).
    \label{eq:t2s}
\end{equation}

In our DualOIE, these two directions have their respective decoders but are combined with a shared encoder. During DualOIE's training, the overall loss is defined as:
\begin{equation}
    \mathcal{L}=\mathcal{L}_{S\rightarrow T}+\mathcal{L}_{T\rightarrow S},
\end{equation}
where $\mathcal{L}_{S\rightarrow T}$ and $\mathcal{L}_{T\rightarrow S}$ are the loss of \textbf{S to T} and \textbf{T to S} (see Section \ref{sec:loss} for details), respectively.

\section{Methodology}

In this section, we describe our DualOIE's details for the two task directions. The overall structure of DualOIE is depicted in Fig. \ref{fig:framework}.

\subsection{Sentence to Triplets}
As mentioned before, this direction is the primary task of our model which aims to generate triplets $Y$ based on the input sentence $x$. We introduce the two steps of this direction in turn as follows.
Specifically, the triplet extraction is divided into two steps:
1) extract the predicate sequence from the input. 
2) use the predicate sequence to induce the generation of triplets.

\subsubsection{Predicate Extraction}
This first step aims to extract a sequence of predicates $y_P$ from the sentence $x$. Predicate extraction is designed upon a Transformer encoder of $L$ blocks, along with a predicate decoder which is also a Transformer decoder.

Formally, we first concatenate the input sentence $x$  with two special tokens to construct this step's input as:
\begin{equation}
    \label{eq:xp}
    x_P=[<\text{sen}>,w_1,...,w_n,<\text{/sen}>].
\end{equation}
Based on the encoder, we calculate the hidden state of $x_P$ as:
\begin{equation}
    \mathbf{H}_P=\text{Encoder}(x_P).
\end{equation}
Then, we use the predicate decoder to generate the predicate sequence $y_P$ in an auto-regressive way:
\begin{equation}    y_i,\mathbf{h}_i^d=\text{Decoder}_P([\mathbf{H}_P;\mathbf{h}_1^d,...,\mathbf{h}_{i-1}^d]),
    \label{eq:decode}
\end{equation}
where $y_i$ is the $i$-th token in $y_P$ and $\mathbf{h}_i^d$ is its decoder state.
Please note that the extracted predicate sequence $y_P$ might consist of multiple predicates, each of which is denoted as $p_i$ and corresponds to one triplet. Thus, we use <rel> and </rel> to split them as below:
\begin{equation}
    y_P=[\text{<rel>},p_1,\text{</rel>},...,\text{<rel>},p_{|Y|},\text{</rel>}].
\end{equation}

Finally, the loss function of predicate extraction is defined as:
\begin{equation}\label{eq:LP}
    \mathcal{L}_P=\sum_{(x_P,y_P)\in \Omega_P}-\log p(y_P|x_P;\theta^e,\theta^d_P),
\end{equation}
where $\Omega_P$ is the training set of all sentence-predicates pairs, $\theta^e$ and $\theta^d_P$ are the parameters of the encoder and the predicate decoder, respectively.

\subsubsection{Triplet Extraction}
This step aims to generate the triplets sequence $Y$ based on $x_P$ and $y_P$. Similar to the previous step, the triplet extraction also consists of an encoder and a triplet decoder. Note that the encoder is shared by the predicate extraction as well.

Formally, given $x_P$ and $y_P$ from the former step, we take $y_P$ as a prompt and concatenate it with $x_P$, to construct the input $x_T$ of this step as:
\begin{equation}
    x_T=[y_P;x_P].
\end{equation}
Then, we use the shared encoder to get $x_T$'s hidden state $\mathbf{H}_T$, and decode the target sequence $y_T$ based on $\mathbf{H}_T$ in the same way of Eq. \eqref{eq:decode}.

To split each extracted triplet, we utilize three pairs of special tokens to formulate a single triplet $t_i$ in $y_T$ as:

\begin{equation}
\label{eq:t}
        t_i=[\text{<sub>},s_i,\text{</sub>},\text{<rel>},p_i,\text{</rel>},\text{<obj>},o_i,\text{</obj>}].
\end{equation}
Similar to $\mathcal{L}_P$ defined in \ref{eq:LP}, the loss function of the triplet extraction $\mathcal{L}_T$ is defined as:
\begin{equation}\label{eq:LT}
    \mathcal{L}_T=\sum_{(x_T,y_T)\in \Omega_T}-\log p(y_T|x_T;\theta^e,\theta^d_T).
\end{equation}

Note that the predicate prompt $y_P$ enables DualOIE to avoid iterative generation and output all triplets in a single decoding process on the second step. 
In contrast, the number of decoding steps in IMoJIE and Gen2OIE depends on the number of triplets corresponding to the input.
In addition, it prompts the model to generate triplets of the same number as the predicates.

\subsection{Triplets to Sentence}
This direction aims to reconstruct a sentence $x$ from the extracted triplets $Y$, which is composed of an encoder and a text decoder. Note that the encoder is shared by the \textbf{S to T} direction as well. 

We formulate the input of this direction as:
\begin{equation}
    x_S=y_T=[t_1,...,t_{|Y|}].
\end{equation}
Next, we derive the hidden state $\mathbf{H}_S$ of $x_S$ from the shared encoder. Then, we use the text decoder to decode the target sequence $y_S$ in the same way of Eq. \eqref{eq:decode}. And the output of this direction $y_S$ is defined in the same way as $x_P$ in Eq. \eqref{eq:xp}. 

Finally, the loss function is defined as:
\begin{equation}
    \mathcal{L}_S=\sum_{(x_S,y_S)\in \Omega_S}-\log p(y_S|x_S;\theta^e,\theta^d_S).
\end{equation}

Note that the shared encoder would be optimized toward both directions, which enables DualOIE to benefit from the dual task simultaneously. 
In addition, the order of triplets in $x_S$ is the same as $Y$, i.e., the appearance order of the predicate in the sentence.

\section{Model Training and Inference}
In this section, we first introduce the training details of loss functions. Then, we describe the inference process of our model.
\subsection{Joint Training}
\label{sec:loss}
We employ a joint training way to \textbf{S to T} direction and \textbf{T to S}. The loss function of DualOIE is defined as:
\begin{equation}
    \begin{aligned}
&\mathcal{L}=\mathcal{L}_{S\rightarrow T}+\mathcal{L}_{T\rightarrow S},\\
    &\mathcal{L}_{S\rightarrow T}=\alpha\mathcal{L}_P+\beta\mathcal{L}_T,\\
    &\mathcal{L}_{T\rightarrow S}=\gamma\mathcal{L}_S,
    \end{aligned}
\end{equation}
where $\mathcal{L}_{S\rightarrow T}$ and $\mathcal{L}_{T\rightarrow S}$ are the loss of \textbf{S to T} and \textbf{T to S}. $\mathcal{L}_P$, $\mathcal{L}_T$ and $\mathcal{L}_S$ are the loss of the predicate extraction, triplet extraction and sentence reconstruction, respectively. In addition, we also use hyper-parameters $\alpha$, $\beta$ and $\gamma$ to balance different objectives. As a result, the overall loss of DualOIE is
\begin{equation}\label{eq:L}  \mathcal{L}=\alpha\mathcal{L}_P+\beta\mathcal{L}_T+\gamma\mathcal{L}_S. 
\end{equation}

\subsection{Model Inference} 
When our model is trained well by the task of \textbf{S to T} and \textbf{T to S}, we only need to use \textbf{S to T} to solve the triplet extraction task inference phase. 
Finally, we split $y_T$ into the triplets of $Y$ by special tokens. 

\section{Experiments}
In this section, we display the results of our extensive experiments to verify DualOIE's advantage, based on which we also provide a detailed analysis. 


\subsection{Datasets} 
\textbf{CaRB}\cite{bhardwaj-etal-2019-carb} is a crowdsourced English benchmark in OpenIE.
However, it does not contain a training set due to the annotation cost, so we use the automatically annotated training data produced by IMoJIE \cite{kolluru-etal-2020-imojie}. 
\textbf{SAOKE}\cite{sun2018logician} is a large-scale human-annotated Chinese dataset. Note that 11\% triplets in SAOKE are implicit.
The statistics of two datasets are reported in Table \ref{stat}.
\begin{table}[b]
\centering
\caption{The statistics of CaRB and SAOKE.}
\scalebox{.9}{
\begin{tabular}{lcccc}
\toprule
      & \multicolumn{2}{c}{CaRB}                                           & \multicolumn{2}{c}{SAOKE}                                          \\ \midrule
      & \multicolumn{1}{l}{\small \# Sentences} & \multicolumn{1}{l}{\small\# Triplets} & \multicolumn{1}{l}{\small\# Sentences} & \multicolumn{1}{l}{\small\# Triplets} \\ \midrule
Train & 92,650                            & 180,689                          & 28,238                            & 73,232                           \\
Test  & 634                              & 2,715                            & 1,569                             & 4,175                            \\ \bottomrule
\end{tabular}
}
\label{stat}
\end{table}

\subsection{Baselines} 
We compared our model against several recent neural OpenIE models, which could be divided into tagging models and generative models.

\subsubsection{Tagging Models}
\emph{RnnOIE} \cite{stanovsky-etal-2018-supervised} takes predicate head with the input together and outputs the tags indicating the token classes. \emph{SpanOIE} \cite{zhan2020span} uses BiLSTM to derive the representation of a span, and then decoders tags from span representations. \emph{IGL-OIE} \cite{kolluru-etal-2020-openie6} proposes an iterative grid labelling system to predict tag sequences. \emph{MacroIE} \cite{yu2021maximal} builds a fact graph based on token spans, and decodes the graph into fact triplets during the inference process.

\subsubsection{Generative Models} \emph{NOIE} \cite{cui-etal-2018-neural} is built with an encoder-decoder structure based on stacked LSTM, where both the copy mechanism and attention mechanism are applied. We provided an advanced version of NOIE in our comparisons, of which the LSTM encoder is replaced with BERT. \emph{IMoJIE} \cite{kolluru-etal-2020-imojie} is also composed of a BERT encoder and an LSTM decoder, and it leverages an iterative generation mechanism to alleviate redundant extraction. \emph{Gen2OIE} \cite{kolluru2022alignment} is an OpenIE system based on two mT5 models. 
Given that \emph{ChatGPT}
is a powerful large language model (LLM) capable of producing contextually appropriate and logically connected responses, We designed the prompts of 0-shot, 3-shot and Chain-Of-Thought (COT) \cite{wei2022chain} to instruct ChatGPT to achieve extraction task. Note that examples were randomly selected in the 3-shot setting.

\subsection{Evaluation Metrics} 
\label{metric}
\textbf{F1(1-1)} \cite{bhardwaj-etal-2019-carb}, is a token level scorer, which creates a label-prediction matching table and then computes precision and recall between each pair.
The overall precision and recall are computed through one-to-one mapping, where a gold triplet and a prediction could match each other only once.
\textbf{F1} \cite{kolluru-etal-2020-openie6} is a variant of F1(1-1), where the recall is computed through multi-to-one mapping. 

\subsection{Implementation Detail}
 During the training, we set the loss coefficient $\alpha$=0.4, $\beta$=0.2, $\gamma$=0.6. For CaRB, we used T5
 \cite{raffel2020exploring} encoder as the shared encoder, and T5 decoder as the predicate decoder, triplet decoder and text decoder, respectively. For SAOKE, we applied the Chinese T5-pegasus
 \cite{zhuiyit5pegasus}. We trained our model
 using Pytorch on an NVIDIA Tesla V100 GPU with 32 GB dedicated memory. The network parameters were optimized by Adam \cite{kingma2014adam} with a learning rate of 2e-5. The batch size was fixed to 32. The total training time was 5 hours. The final displayed results were reported as the average of the results of 5 random seeds.


\subsection{Overall Comparison Results}
\begin{table}[t]
\centering
\caption{Main results on CaRB and SAOKE.}
\label{main_result}
\tabcolsep=7pt
\scalebox{1}{
\begin{tabular}{lcccc}
\toprule
Model & \multicolumn{2}{c}{CaRB} & \multicolumn{2}{c}{SAOKE} \\
              & F1(1-1)      & F1        & F1(1-1)          & F1               \\ \midrule
SpanOIE        & 37.9         & 48.5      & 38.6    & 41.2 \\
RnnOIE         & 39.5         & 49.0      & 39.9    & 42.7 \\
IGL-OIE        & 41.1         & 52.4      & 42.4    & 44.5 \\
MacroIE        & 43.5         & 54.8      & 43.7    & 45.6 \\ \midrule
NOIE+BERT      & 38.7         & 51.6      & 50.2    & 51.3 \\
IMoJIE         & 41.4         & 53.5      & 52.4    & 54.3 \\
Gen2OIE        & 43.4         & 54.6      & 53.6    & 55.4 \\
ChatGPT (0-shot)& 39.5         & 50.3      & 51.9    & 53.7 \\
ChatGPT (3-shot)& 40.7         & 51.1      & 52.6    & 54.3 \\
ChatGPT (COT)   & 42.2         & 53.4      & 54.0    & 55.9 \\ \midrule
DualOIE w/o D  & 50.3         & 54.2      & 56.4    & 57.7 \\
DualOIE w/o P  & 50.9         & 55.5      & 56.8    & 58.0 \\
DualOIE &\textbf{51.5} &\textbf{56.3} &\textbf{58.1}& \textbf{59.5}            \\ \bottomrule
\end{tabular}
}
\end{table}

\begin{table}[t]
\centering
\caption{A toy example of the ablation study of DualOIE.}
\scalebox{.9}{
        \centering
        \begin{tabular}{ll}
        \toprule
        \textbf{Sentence} &      He is idolized , receiving the name of ``God''.                \\ \midrule
        DualOIE w/o D & (He , is idolized receiving, the name of ``God'')                \\
        DualOIE w/o P & (He , receiving, the name of ``God'')                            \\
        DualOIE       & (He, is, idolized) ;  (He, is receiving,  the name of ``God'') \\ \bottomrule
        \end{tabular}}
        \label{tb:as}
\end{table}

\begin{table}
\centering
\caption{The results of DualOIE on the two datasets with different extraction order in the task of \textbf{S to T}.}
\scalebox{1}{
\begin{tabular}{lcccc}
\toprule
                & \multicolumn{2}{c}{CaRB} & \multicolumn{2}{c}{SAOKE} \\
Models          & F1(1-1)   & F1      & F1(1-1)        & F1     \\ \midrule
DualOIE-S       & 44.8      & 51.4    & 54.3           & 55.6     \\
DualOIE-O       & 43.6      & 50.1    & 51.6           & 53.2     \\ 
DualOIE         & 51.5     & 56.3   & 58.1          & 59.5     \\ 
\bottomrule
\end{tabular}
}
\label{tb:order}
\end{table}

The comparison results between our DualOIE and the baselines are shown in Table \ref{main_result}.
According to the results, we have the following conclusions. 

1) DualOIE achieves the best extraction performance on the two datasets under each metric, fully justifying the effectiveness of both employing the dual task in joint training and the predicate prompt in extraction process.

2) On CaRB, there is always a gap of about 10\% or more between the two metrics in the baselines, which is mainly caused by the low quality of automatically-derived training data. DualOIE decreases this gap to 4.8\%, indicating that a triplet well-predicted by it is less mapped to several different gold triplets. In other words, the output of DualOIE is much more precise and robust in one-to-one mapping metric.

3) On SAOKE, DualOIE outperforms the best baseline by a large margin.
In addition, note that generative models perform better overall than tagging models on SAOKE. This is because the implicit triplets in SAOKE were missed by the tagging models due to their inability to introduce new words.
In addition, the gap between the two metrics is much smaller and more stable due to the high quality of SAOKE. 

\subsection{In-depth Investigations}

\begin{figure}[t]
    \centering
		\centering
		\includegraphics[width=.8\linewidth]{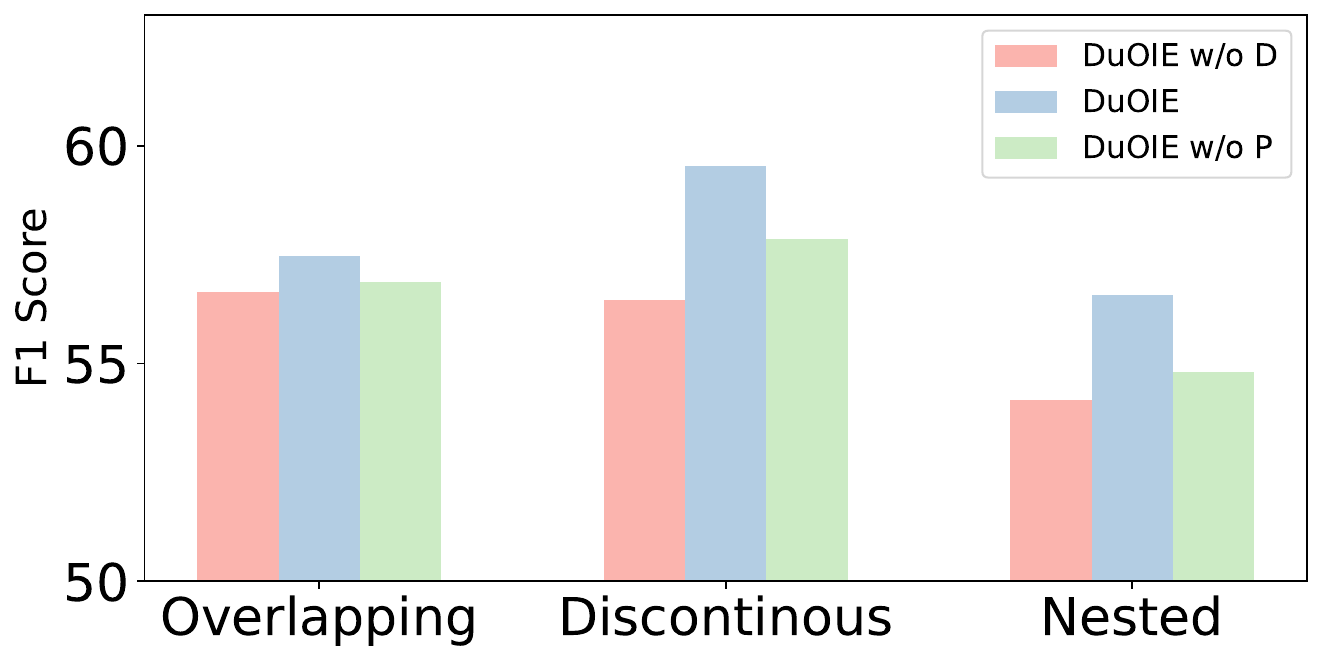}
  \vspace{-0.2cm}
		\caption{Performance on extraction of complicated triplets in SAOKE.}
		\label{fig:ab}
\end{figure}

\subsubsection{Ablation Study}
To evaluate the impact of each component in DualOIE, we compared it with two ablated variants: \emph{DualOIE w/o D} whose dual task is removed, and \emph{DualOIE w/o P} in which the predicate prompt is removed and the triplets are directly generated by the decoder. 

From the comparison results shown in Table \ref{main_result}, we find: 
1) The two ablated variants still have an advantage over most baselines, justifying the effectiveness of incorporating either the predicate prompt or the dual task. 
2) The removal of the dual task leads to an apparent performance drop on both datasets, i.e., (1.2\%, 2.1\%) on CaRB and (1.7\%, 1.8\%) on SAOKE. This proves the importance of the dual task for enhanced extraction. And the removal of the predicate prompt also decreases our model's performance.

Fig. \ref{fig:ab} shows the comparison results on complicated cases, which indicates that the introduction of duality could improve performance on these challenges. Since the syntactic knowledge endowed by the dual task is helpful to solve discontinuous and nested triplets, which arise from complicated sentence structure. In addition, the dual task encourages the model to capture diverse relations between arguments, which benefits the solving of overlapping. 
We also give a case example among ablated versions. 
As shown in Table \ref{tb:as}, DualOIE correctly outputted both triplets.
The removal of the duality prevented the model from effectively learning the structure of the sentence, resulting in confusion between the two triplets and outputting an incorrect triplet. 
Additionally, removing the predicate prompt caused the model to miss one of the triplets.

\begin{table*}[t]
\centering
\caption{Tuning results of loss coefficients on SAOKE.}
\label{tb:loss}
\tabcolsep=13pt 
\scalebox{1}{ 
\begin{tabular}{|c|cc|cc|cc|}
\hline
               & \multicolumn{2}{c|}{$\alpha+\beta<\gamma$} & \multicolumn{2}{c|}{$\alpha+\beta=\gamma$} & \multicolumn{2}{c|}{$\alpha+\beta>\gamma$} \\ 
               & ($\alpha$, $\beta$, $\gamma$)     & F1       & ($\alpha$, $\beta$, $\gamma$)     & F1       & ($\alpha$, $\beta$, $\gamma$)     & F1      \\ \hline
               & (0.2,0.4,1.2)                   & 56.4     & (0.2,0.4,0.6)                   & 57.1     & (0.2,0.4,0.3)                   & 55.9    \\
$\alpha<\beta$ & (0.2,0.6,1.6)                   & 56.3     & (0.2,0.6,0.8)                   & 57.5     & (0.2,0.6,0.4)                   & 54.9    \\
               & (0.2,0.8,2.0)                   & 54.0     & (0.2,0.8,1.0)                   & 55.6     & (0.2,0.8,0.5)                   & 53.3    \\ \hline
               & (0.2,0.2,0.8)                   & 55.7     & (0.2,0.2,0.4)                   & 56.5     & (0.2,0.2,0.2)                   & 55.3    \\
$\alpha=\beta$ & (0.4,0.4,1.6)                   & 57.0     & (0.4,0.4,0.8)                   & 57.3     & (0.4,0.4,0.4)                   & 55.8    \\
               & (0.6,0.6,2.4)                   & 55.7     & (0.6,0.6,1.2)                   & 57.0     & (0.6,0.6,0.6)                   & 54.9    \\ \hline
               & (0.4,0.2,1.2)                   & 58.1     & (0.4,0.2,0.6)                   & \textbf{59.5}     & (0.4,0.2,0.3)                   & 56.2    \\
$\alpha>\beta$ & (0.6,0.2,1.6)                   & 57.2     & (0.6,0.2,0.8)                   & 58.4     & (0.6,0.2,0.4)                   & 56.0    \\
               & (0.8,0.2,2.0)                   & 56.7     & (0.8,0.2,1.0)                   & 57.3     & (0.8,0.2,0.5)                   & 55.4    \\ \hline
\end{tabular}
}
\end{table*}

\begin{figure}[t]
    \centering
		\includegraphics[width=\linewidth]{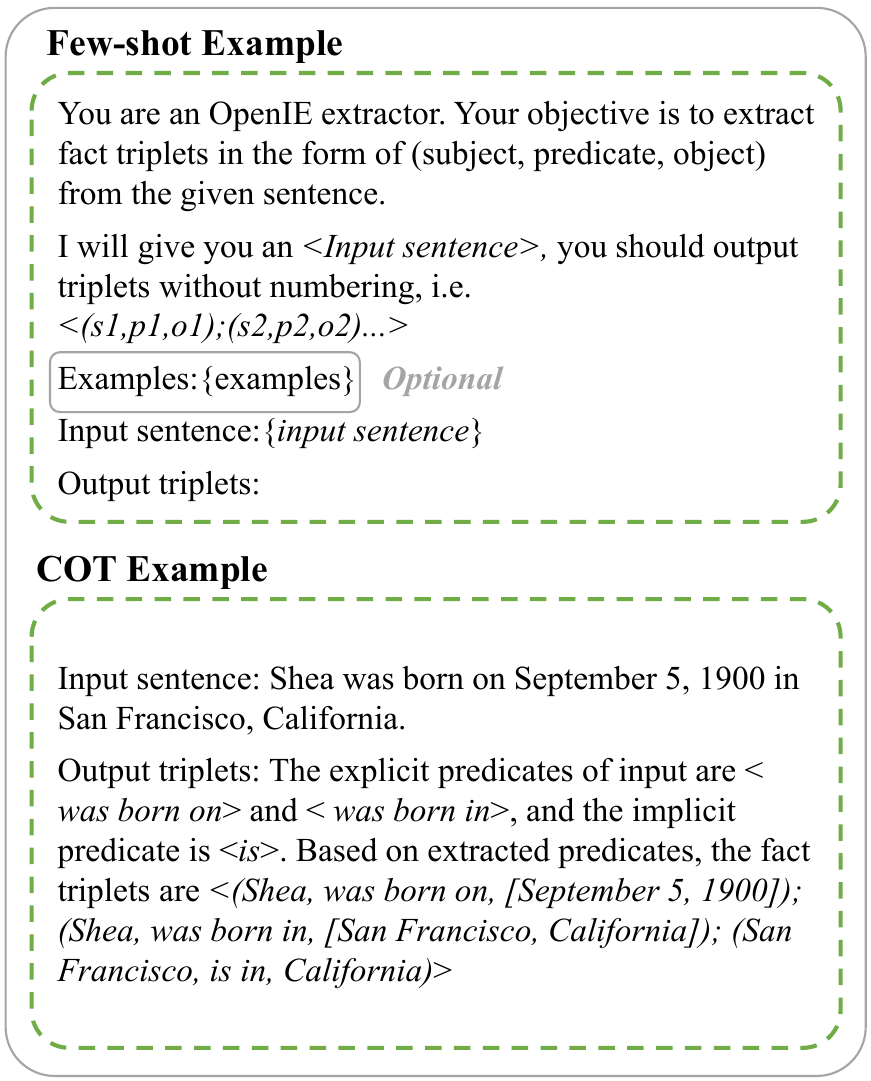}
  \caption{The prompts we designed to instruct ChatGPT to perform the OpenIE task.}
		\label{fig:gpt_prompt}
\end{figure}

\subsubsection{Analysis on Extraction Order}

We also conducted experiments to investigate the effect of taking different terms in a triplet as the prompt. Table \ref{tb:order} details the performance of various prompt settings, where DualOIE-S(O) is the version where all subjects(objects) are extracted at first as the prompt, and then concatenated with the input to generate the triplets. From the table, we observe that the predicate prompt performs best followed by the subject prompt, and the object prompt performs worst. 
It is mainly due to the fact that a subject or object could appear in multiple triplets, and sometimes a subject in one triplet might even be an object in another triplet. Such complicated situations would confuse the model during triplets generation, while the predicate prompt could reduce this uncertainty since an extracted predicate corresponds to a certain triplet.

\subsubsection{Impact of Loss Coefficient}

As formalized in Eq. \ref{eq:L}, the loss coefficient $\alpha$ and $\beta$ indicate importance of the primary extraction task \textbf{S to T}, while $\gamma$ indicates the importance of the dual task \textbf{T to S}. Thus, to analyze the importance of two task directions, we compared $\alpha+\beta$ with $\gamma$ where $\frac{\gamma}{\alpha+\beta}$ was set to 0.5, 1 and 2, respectively. In addition, to analyze the importance of the two steps in \textbf{S to T}, we also compared $\alpha$ with $\beta$ where the value ratio between them was set to 2, 3 and 4, respectively.

From Table \ref{tb:loss}, we have the following conclusions. 1) The case of $\alpha+\beta=\gamma$ performs better, indicating that the dual task is as important as the primary task, but the balance between the two tasks should be kept. 2) the case of $\alpha>\beta$ shows its advantage, indicating that the predicate extraction is tougher and thus requires more concentrations. Besides, more concentrations on the predicate extraction could alleviate the cascading error in DualOIE's pipeline-based extraction framework.

\subsubsection{Analysis on ChatGPT's Performance}

We have examined the OpenIE performance of ChatGPT in three prompt settings: 0-shot, 3-shot and COT. 
Fig. \ref{fig:gpt_prompt} shows the main prompt and the COT example. 
Table \ref{main_result}'s results indicate that ChatGPT is still competitive with fine-tuned baselines even in the 0-shot scenario. 
Compared with 3-shot prompt, COT prompt is more effective since it enables ChatGPT to learn the knack of extracting the predicate followed by extracting the subject and object, which is in fact an effective extraction manner adopted in our DualOIE.
{
However, ChatGPT still shows less efficacy than DualOIE, since it sometimes fails to correctly comprehend the relations between arguments and tends to confuse the boundaries of arguments \cite{han2023information}. For example, given the input in Table \ref{tb:as}, the output of ChatGPT(COT) is (He, idolized, receiving the name of “God”), because it misjudges the predicate between arguments.
} 

In addition, we observe that ChatGPT sometimes generates triples that do not align with the input sentence, also possibly due to hallucination \cite{bang2023multitask}.

\subsubsection{Impact of Triplet Number}
\begin{figure}[t]
		\centering
		\includegraphics[width=.85\linewidth]{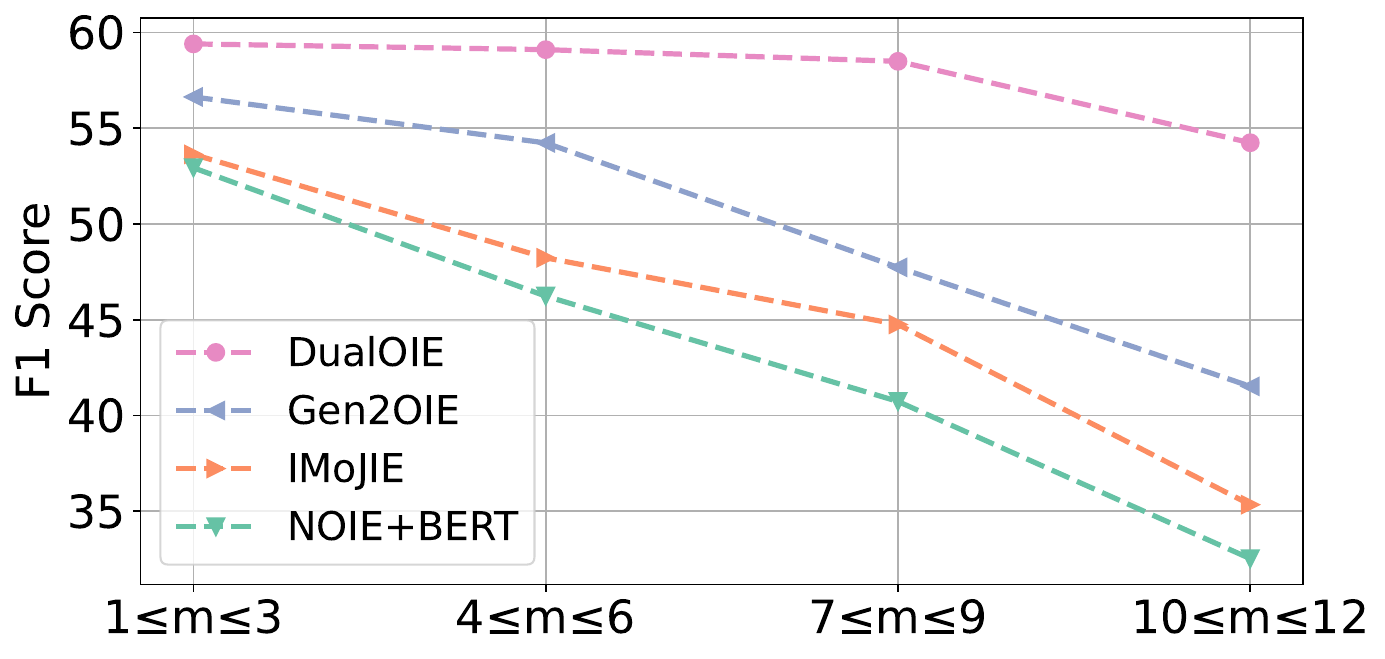}
		\caption{Performance comparisons on the groups with different triplet numbers ($m$) of a sentence in SAOKE.}
		\label{fig:last}
\end{figure}

As illustrated in Fig. \ref{fig:last}, it is harder for the extraction models to obtain satisfactory performance when the sentence's structure is getting more complicated, where the number of potential triplets increases.
To further investigate the models' capabilities of handling the convoluted sentences with multiple triplets, we divided the samples (sentences) in SAOKE into 4 groups according to the triplet number in one sentence, and then evaluated all compared models in each group. Fig. \ref{fig:last} displays the relevant results where $m$ is the triplet number. It shows that DualOIE outperforms baselines on all groups by a large margin. Particularly, all baselines' performance degrades sharply when more triplets exist in a sentence. Comparatively, DualOIE's performance drop is slighter, justifying its stability of extracting triplets in complicated situations. 

\begin{figure}[t]
        \captionof{table}{Pred-F1 and Trip-F1 are defined identically to those in Sec.\ref{metric}. Gold refers to the situation where golden predicate prompt is provided.}
        \scalebox{1}{
        \tabcolsep=4.5pt

        \begin{tabular}{lccc}
        \toprule
        Model   & Dataset & Pred-F1 & Trip-F1 \\ \midrule
        DualOIE & CaRB    & 65.5         & 56.3       \\
        DualOIE & CaRB    & Gold         & 63.2       \\ 
        \midrule
        DualOIE & SAOKE   & 65.7         & 59.5       \\
        DualOIE & SAOKE   & Gold         & 65.6       \\ 
        \bottomrule
        \end{tabular}}
        \label{tb:pe}
\end{figure}

\subsubsection{Correlation of the Two Tasks}
\begin{figure}[t]
    \centering
		\centering
		\setlength{\abovecaptionskip}{0.15cm}
		\includegraphics[width=0.85\linewidth]{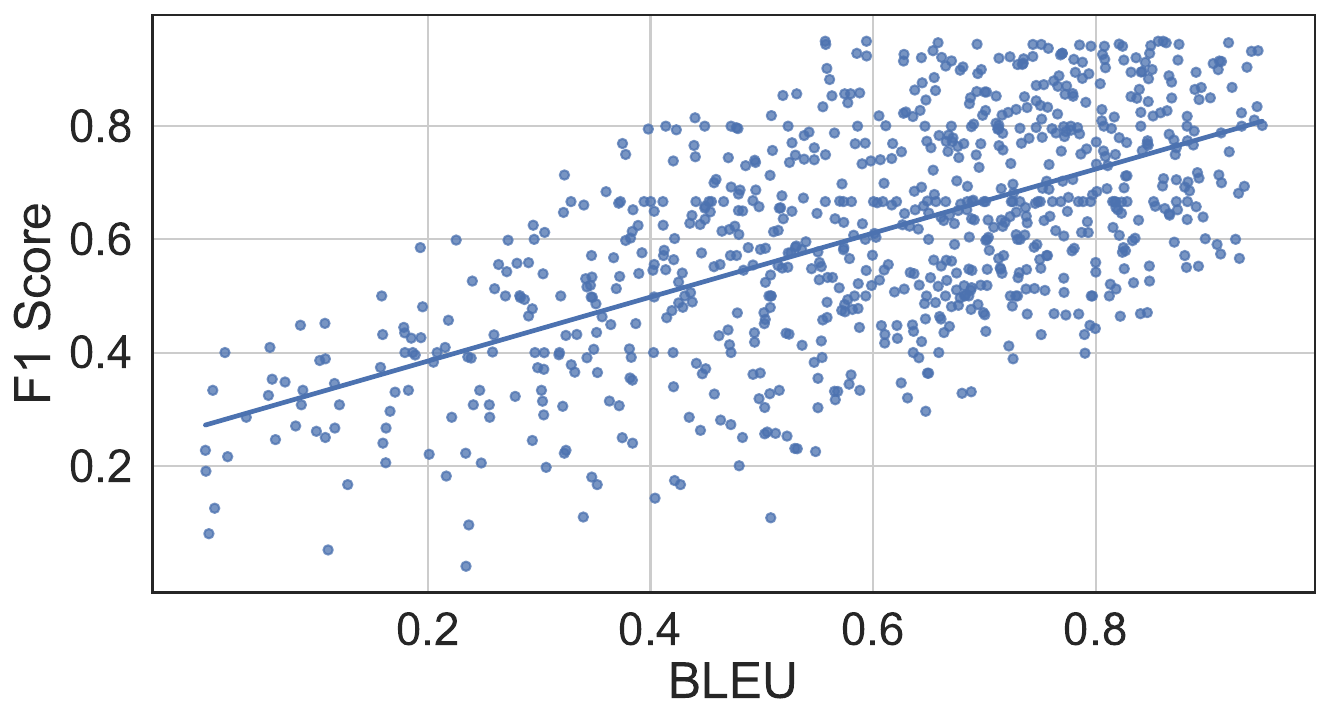}
  \caption{Correlation analysis between the BLEU of \textbf{T to S} (X-axis) and the F1 of \textbf{S to T} (Y-axis).}
		\label{fig:dual_task}
\end{figure}

We have also observed a phenomenon of mutual promotion between the dual task and the primary task (triplet extraction).
We used BLEU \cite{papineni2002bleu} to evaluate the quality of the restored sentences generated by \textbf{T to S}, and compared it with the F1 of \textbf{S to T}. 
The BLEU score varies between 0 and 1, showing how similar the generated text is to the gold text. 
The scatter and the regression line in Fig. \ref{fig:dual_task} show the positive correlation between BLEU and F1, implying that two tasks could improve the performance of each other.

\subsubsection{Analysis on Cascading Error} 
Since we used a pipeline-style extraction structure in the \textbf{T to S} direction, it may cause the cascading error. Therefore, we designed an experiment to analyze the impact of the cascading error on final triplet extraction. As shown in Table \ref{tb:pe}, the F1 of the predicate extraction on two datasets are 65.5\% and 65.7\%. Given the prompt with golden predicates, there is 6.9\% and 6.1\% improvement in extraction results.

\begin{CJK}{UTF8}{gbsn}
\section{Construction of MTOIE }
MTOIE is a large-scale Chinese dataset and mainly focuses on noun-based attributes of POIs, which refer to the predicates in user comments, since the noun-based attributes often reflect customer attitudes to the POIs.
We randomly collected sentences from user comments of POIs on the Meituan platform, covering domains like food, entertainment and accommodation.
Then we sent these sentences to the annotators and asked them to annotate triples in the form of (\emph{entity}, \emph{attribute}, \emph{attribute-value}). 
The annotation for MTOIE is not easy for those inexperienced workers, especially for triples with implicit attributes. Thus, we employed 12 professional annotators served in Meituan. To obtain a high-quality dataset, we first trained them to annotate the triples with explicit attributes, and then evaluated their performance. 
Only the annotators achieving high evaluation scores were allowed to annotate the triplets with implicit attributes.

\subsection{Explicit Triplet Annotation}
The triplets with explicit attributes require the annotators to tag out the entity, attribute, and attribute-value in the sentence in order. 
To make the annotators quickly understand the definition of attributes, we leveraged a predefined attribute vocabulary\footnote{About 2,000 attributes accumulated by the Meituan company.} to match the attributes in the sentence and marked them to indicate possible attributes. 
Following the philosophy of previous OpenIE benchmarks \cite{bhardwaj-etal-2019-carb,sun2018logician}, we identified the following three guidelines for MTOIE annotation. \\

\noindent\textbf{Completeness} We aim to extract all triplets with marked attributes from the sentence. Firstly, the annotators must carefully examine all marked attributes, and annotate the entities and attribute-values corresponding to the attributes that were judged as correct. Then, the annotators checked again whether some attributes are missing from the marked attributes. \\
\noindent\textbf{Assertedness} The annotation for the triplets with  explicit attributes follows the principle of ``literally honest'', i.e., entity, attribute and attribute-value must be asserted by the original sentence. \\
\noindent\textbf{Atomicity} Each triplet is required to be an indivisible unit. The annotators must extract multiple atomic triplets from the sentence that has conjunctions. For example, given a sentence ``面包有两种口味，菠萝味和芒果味。/Bread has two flavors, pineapple and mango.'', the annotators should annotate (面包/Bread, 口味/flavors, 菠萝味/pineapple) and (面包/Bread, 口味/flavors, 芒果味/mango) instead of (面包/Bread, 口味/flavors, 菠萝味和芒果味/pineapple and mango).

\subsection{Implicit Triplet Annotation}
Implicit attributes are the most challenging part of the annotation. 
Manually annotating implicit attributes from the unstructured text tends to cause missing and incorrect annotations.
Therefore, we proposed an iterative annotation framework to reduce the annotation difficulty, simplifying the ``manual inference'' to ``model prediction along with by manual discrimination'' (as shown in Algorithm \ref{code1}).

\begin{algorithm}[t]
  \caption{Iterative Annotation Framework.}
  \label{code1}
  \begin{algorithmic}[1]
    \Require
      $D_{ex}$: annotated explicit samples;
      $D_{un}$: unannotated sentences.
    \Ensure
      $D_{im}$: annotated implicit samples;
    \State Mask the attribute in the sentence of $D_{ex}$ to get the pseudo implicit samples $D_{ps}$;
    \For{$k = 1 \to K$}
        \State Train OAE model $M_k$ with $D_{ps}$;
        \State Use model $M_k$ to predict $D_{un}$ and obtain semi-supervised samples;
        \State Deliver semi-supervised samples to the annotators, get annotated implicit samples $D_{im}$;
        \State Update $D_{ps}$ with $D_{im}$;
    \EndFor
  \end{algorithmic}
\end{algorithm}

At first, we masked the attributes in the explicit samples $D_{ex}$ to obtain the pseudo implicit samples $D_{ps}$. For example, given ``这家游泳池水深2米。/The swimming pool is 2 meters deep'', the original explicit triplet is (游泳池/The swimming pool, 水深/deep, 2米/2 meters), then we removed the attribute ``水深/deep'' from the sentence, and thus the triplet becomes a pseudo implicit one. 

Second, we trained the model $M$ based on $D_{ps}$ and then used $M$ to predict the unannotated sentences $D_{un}$ and obtain the semi-supervised samples. Then, the samples whose attributes do not appear in the sentences were delivered to the annotators. This method greatly reduces the annotation difficulty, and the annotators only need to judge whether the implicit triplets could be inferred from the sentence.

Third, the annotated implicit samples were used to train the model $M$ continuously. After $K$ rounds of iteration, we obtained the final annotated implicit dataset. Following \cite{ding2021few}, we calculated the Cohen’s Kappa to measure the agreements between annotators, and the Kappa score is 88.36\%.

\subsection{Dataset Statistics}
In MTOIE, we have collected 54,060 Chinese sentences and 87,971 triplets in total. Among them, 15,137(28\%) sentences contain implicit triplets, 29,290(33\%) triplets are implicit, and there are 2109 types of implicit predicates.
To estimate the overall quality of the dataset, two volunteers randomly picked 200 sentences from the dataset and evaluated them carefully.
The result shows that the triple-level precision and recall are 94.6\% and 85.5\%, respectively.
\end{CJK}

\section{Application on Meituan}
We have also evaluated our proposed DualOIE on our built MTOIE and verified its effectiveness for the real scenario, i.e., the search service on the Meituan Platform. 

\paragraph{Metric} Our concentration on POI attributes makes the triplets in MTOIE much shorter. Thus using the token-level metrics might overestimate model performance. Therefore, we adopted a more strict matching strategy, where each subject and object are evaluated by a full match. 
Besides, we used a semantic similarity model\footnote{https://github.com/shibing624/text2vec} to determine whether the predicted attribute and the gold have the same meaning. The threshold is set to 0.7.

\paragraph{Comparison Results}
\begin{table*}[t]
\centering
\caption{Triplet extraction comparisons of different models on MTOIE.}\label{tb:MTOIE}
\vspace{-0.2cm}
\scalebox{.9}{
\begin{tabular}{|l|cccccc|cccccc|c|}
\Xhline{1pt}
          & \multicolumn{6}{c|}{Implicit Triplets}                             & \multicolumn{6}{c|}{All Triplets}                                  & \multicolumn{1}{l|}{Speed} \\ \cline{2-13}
Model    & Precision     &      & Recall        &      & F1            &      & Precision     &      & Recall        &      & F1            &      & sen.\#/sec                   \\ 
\Xhline{1pt}
NOIE+BERT & 60.5 & $\uparrow$36\% & 51.5& $\uparrow$37\% & 55.6& $\uparrow$36\% & 64.8 & $\uparrow$32\% & 54.7& $\uparrow$32\% & 59.3 & $\uparrow$32\% & \textbf{11.5}             \\
IMoJIE    & 61.7 & $\uparrow$33\% & 54.8& $\uparrow$29\% & 58.0& $\uparrow$31\% & 67.2 & $\uparrow$27\% & 61.0& $\uparrow$19\% & 64.0 & $\uparrow$22\% & 2.6                       \\
Gen2OIE   & 69.1 & $\uparrow$19\% & 59.7& $\uparrow$18\% & 64.1& $\uparrow$18\% & 73.4 & $\uparrow$16\% & 63.7& $\uparrow$14\% & 68.2 & $\uparrow$15\% & 2.8                       \\ \hline
DualOIE   & \textbf{82.1} &      & \textbf{70.5} &      & \textbf{75.9} &      & \textbf{85.3} &      & \textbf{72.3} &      & \textbf{78.3} &      & 3.7                       \\ 
\Xhline{1pt}
\end{tabular}
}
\end{table*}

We compared DualOIE's extraction performance on MTOIE with NOIE+BERT, IMoJIE and Gen2OIE.
{ChatGPT was not included as MTOIE has not been released yet. Using ChatGPT could lead to data leaks. The relevant results will be updated on GitHub once it is released in future.}
The results in Table \ref{tb:MTOIE} justify our DualOIE's advantage on both implicit triplets and all triplets under strict matching metrics. 
{It shows that the relative performance improvement of DualOIE over the baselines on implicit triplets is more significant than that on all triplets, demonstrating DualOIE's stronger capability of extracting implicit information.}
We also compared all the models' efficiency through analyzing the number of sentences that the models can process per second, as shown in the last column of Table \ref{tb:MTOIE}. As mentioned before, DualOIE shows an advantage over IMoJIE and Gen2OIE due to its decoding steps independent of the triplets number, while the two competitors spend more time on sentences with multiple triplets. 
{Note that the single-turn decoding model NOIE+BERT is much faster than the other three generative methods, because it adopts beam search to output multiple triplets instead of a sequence of all triplets. Since the number of beams should be pre-decided, it fails to naturally adapt the extraction number to the input sentence.}

\begin{CJK}{UTF8}{gbsn}
\begin{table*}[!htb]
\centering
\caption{Extraction results for two cases where the incorrect terms are marked red.}
\vspace{-0.2cm}
\scalebox{.9}{
\begin{tabular}{ll}
\toprule
Sentence & 现在许多博物馆，纪念馆票价都免费了。/Nowadays many museums, monuments tickets are free. \\ \midrule
IMoJIE   & (纪念馆/monuments，票价/tickets，免费/free)             \\
Gen2OIE & (纪念馆/monuments，票价/tickets，免费/free)             \\
DualOIE  & (博物馆/museums，票价/tickets，免费/free), (纪念馆/monuments，票价/tickets，免费/free) \\ 
\toprule
Sentence & 帮我做美睫的小姐姐很专业，效果非常满意。\\
 & The lady helping me with beautiful eyelashes is professional, the effect is very satisfactory.  \\
\midrule
IMoJIE & (\textcolor{red}{小姐姐/The lady}，效果/effect，满意/satisfactory)             \\
Gen2OIE  & (美睫/beautiful eyelashes，效果/effect，满意/satisfactory), (\textcolor{red}{美睫/beautiful eyelashes}，手法/skill，专业/professional)  \\
DualOIE & (美睫/beautiful eyelashes，效果/effect，满意/satisfactory), (小姐姐/The lady，手法/skill，专业/professional)  \\
\bottomrule
\end{tabular}}
\label{tb:case}
\end{table*}
\end{CJK}

Furthermore, we display the extraction results of our DualOIE and two state-of-the-art baselines, i.e., IMoJIE and Gen2OIE, on some difficult cases. As shown in Table \ref{tb:case}, for the first case, IMoJIE and Gen2OIE both miss a gold triplet.  
For the second case, IMoJIE predicts a wrong triplet and misses an implicit triplet, and Gen2OIE predicts the wrong subject of the implicit triplet. The two baselines' inferior results are due to their insufficient recognition of the input sentence's structure. By contrast, DualOIE generates more correct triplets with the help of the dual task.

\paragraph{A/B Test}
\begin{CJK}{UTF8}{gbsn}
We also conducted an online A/B test to show how our model helps the Meituan platform improve search performance. 
Specifically, given a POI, we first used DualOIE to extract a collection of triplets from every user comment of it.
Then, we used the semantic similarity model to normalize the collection as follows. 
1) The similarity scores of the subject, attribute (predicate) and object were computed between two triplets, and they would be considered expressing the same fact if their scores are over (0.7, 0.7, 0.7).
2) The triplets with the same meaning were replaced by the most frequent one in the collection, and the reserved triplets were ranked by their frequencies.
Finally, given a triplet such as (海底捞/Haidilao，食材/food，新鲜/fresh) in the normalized collection, we concatenated the attribute ``食材/food'' with the object ``新鲜/fresh'' as the given POI's label ``食材新鲜/food fresh''.

For the online A/B test, we used two buckets where each bucket containing 25\% randomly selected users. For one bucket, the system returned the search results using the labels to filter out POIs.
For another bucket, the system directly returned the POIs by its default principle.
We ran our A/B test on Meituan's searching system and compared daily average performance on query view click-through rate (QV-CTR) and unique visitor click-through rate (UV-CTR), which refer to the ratio of clicked queries to all queries and the ratio of clicked queries to all unique user-query pairs, respectively.
After the running of 15 days, we noticed that QV-CTR and UV-CTR increased by 0.93\% and 0.56\%, respectively. These results indicate that the POI labels obtained through DualOIE indeed improve the performance of Meituan's search service.
\end{CJK}
\section{Related Work}
Traditional OpenIE systems such as
TextRunner \cite{2007Open}, ReVerb \cite{fader2011identifying}, OLLIE \cite{schmitz2012open}, Stanford-IE \cite{angeli2015leveraging}, OpenIE-5 \cite{saha2017bootstrapping} and MinIE \cite{gashteovski2017minie} are mainly based on rules and statistics, which are combined with some modules like POS taggers, SRL parsers and chunkers.

In recent years, researchers mainly focus on neural solutions, which could be divided into two categories: tagging-based systems and generative systems. Tagging-based systems model OpenIE as a sequence labelling problem. For example, RnnOIE \cite{stanovsky-etal-2018-supervised} introduces a custom BIO tagging scheme and takes the predicate head with the input together.
SenseOIE \cite{roy2019supervising} introduces the dependency tree, where the feature of a token is influenced by its parent and children. Besides, it also uses lexical and syntactic features such as POS embedding and syntactic role embedding.
OpenIE6 \cite{kolluru-etal-2020-openie6} is built as an iterative grid labelling (IGL) system and achieves predictions in a 2D grid labelling way. 
SpanOIE \cite{zhan2020span} uses BiLSTM to derive the representation of a span, from which the tag of each token is predicted.
In MacroIE \cite{yu2021maximal}, a fact graph is built based on the token spans, and decoded into fact triplets during the inference process.
On the other hand, generative systems model OpenIE as a Seq2Seq problem, and thus are capable of introducing new words to deal with implicit triplets.
NOIE \cite{cui-etal-2018-neural} uses LSTM as both encoder and decoder, and introduces copy attention to address the out-of-vocabulary problem. 
In Logician \cite{sun2018logician}, BiGRU is used as both encoder and decoder. As well, the coverage mechanism and gated dependency mechanism are both introduced. 
IMoJIE \cite{kolluru-etal-2020-imojie} uses BERT as encoder and LSTM as decoder. It leverages an iterative generation mechanism to alleviate the redundancy in triplets generation. 
Gen2OIE \cite{kolluru2022alignment} is composed of two mT5 \cite{xue2020mt5} models, which reconstructs multiple inputs based on the first model, and then uses the second model to generate triplets. 
\vspace{-.5cm}

\section{Conclusion}
We propose a novel generative OpenIE model DualOIE based on an auxiliary dual task with predicate prompt. By exploiting duality, DualOIE performs better than the previous OpenIE models, since it can better understand the sentence structure and capture diverse relations between arguments. We also provide a high-quality OpenIE dataset MTOIE, which is constructed from the user comments on Meituan's platform. The experiments on public benchmarks and MTOIE justify DualOIE's superiority over the baselines. The online A/B test on Meituan platform also suggests the extracted triplets can promote the real-world search service.


\section{ACKNOWLEDGEMENT}
This paper was supported by Chinese NSF Major Research Plan No.92270121, NSF funding No.62306112, Shanghai Sailing Program No.23YF1409400, Shanghai Science and Technology Innovation Action Plan No.21511100401.
\clearpage
\bibliographystyle{ACM-Reference-Format}
\bibliography{reference}


\begin{thebibliography}{29}


\ifx \showCODEN    \undefined \def \showCODEN     #1{\unskip}     \fi
\ifx \showDOI      \undefined \def \showDOI       #1{#1}\fi
\ifx \showISBNx    \undefined \def \showISBNx     #1{\unskip}     \fi
\ifx \showISBNxiii \undefined \def \showISBNxiii  #1{\unskip}     \fi
\ifx \showISSN     \undefined \def \showISSN      #1{\unskip}     \fi
\ifx \showLCCN     \undefined \def \showLCCN      #1{\unskip}     \fi
\ifx \shownote     \undefined \def \shownote      #1{#1}          \fi
\ifx \showarticletitle \undefined \def \showarticletitle #1{#1}   \fi
\ifx \showURL      \undefined \def \showURL       {\relax}        \fi
\providecommand\bibfield[2]{#2}
\providecommand\bibinfo[2]{#2}
\providecommand\natexlab[1]{#1}
\providecommand\showeprint[2][]{arXiv:#2}

\bibitem[Angeli et~al\mbox{.}(2015)]%
        {angeli2015leveraging}
\bibfield{author}{\bibinfo{person}{Gabor Angeli}, \bibinfo{person}{Melvin Jose~Johnson Premkumar}, {and} \bibinfo{person}{Christopher~D Manning}.} \bibinfo{year}{2015}\natexlab{}.
\newblock \showarticletitle{Leveraging linguistic structure for open domain information extraction}. In \bibinfo{booktitle}{\emph{Proceedings of the 53rd Annual Meeting of the Association for Computational Linguistics and the 7th International Joint Conference on Natural Language Processing (Volume 1: Long Papers)}}. \bibinfo{pages}{344--354}.
\newblock


\bibitem[Bang et~al\mbox{.}(2023)]%
        {bang2023multitask}
\bibfield{author}{\bibinfo{person}{Yejin Bang}, \bibinfo{person}{Samuel Cahyawijaya}, \bibinfo{person}{Nayeon Lee}, \bibinfo{person}{Wenliang Dai}, \bibinfo{person}{Dan Su}, \bibinfo{person}{Bryan Wilie}, \bibinfo{person}{Holy Lovenia}, \bibinfo{person}{Ziwei Ji}, \bibinfo{person}{Tiezheng Yu}, \bibinfo{person}{Willy Chung}, {et~al\mbox{.}}} \bibinfo{year}{2023}\natexlab{}.
\newblock \showarticletitle{A multitask, multilingual, multimodal evaluation of chatgpt on reasoning, hallucination, and interactivity}.
\newblock \bibinfo{journal}{\emph{arXiv preprint arXiv:2302.04023}} (\bibinfo{year}{2023}).
\newblock


\bibitem[Banko et~al\mbox{.}(2007)]%
        {2007Open}
\bibfield{author}{\bibinfo{person}{M. Banko}, \bibinfo{person}{M.~J. Cafarella}, \bibinfo{person}{S. Soderland}, \bibinfo{person}{M. Broadhead}, {and} \bibinfo{person}{O. Etzioni}.} \bibinfo{year}{2007}\natexlab{}.
\newblock \showarticletitle{Open information extraction from the web}.
\newblock \bibinfo{journal}{\emph{Communications of the Acm}} (\bibinfo{year}{2007}).
\newblock


\bibitem[Bhardwaj et~al\mbox{.}(2019)]%
        {bhardwaj-etal-2019-carb}
\bibfield{author}{\bibinfo{person}{Sangnie Bhardwaj}, \bibinfo{person}{Samarth Aggarwal}, {and} \bibinfo{person}{Mausam Mausam}.} \bibinfo{year}{2019}\natexlab{}.
\newblock \showarticletitle{{C}a{RB}: A Crowdsourced Benchmark for Open {IE}}. In \bibinfo{booktitle}{\emph{Proceedings of the 2019 Conference on Empirical Methods in Natural Language Processing and the 9th International Joint Conference on Natural Language Processing (EMNLP-IJCNLP)}}. \bibinfo{publisher}{Association for Computational Linguistics}, \bibinfo{address}{Hong Kong, China}, \bibinfo{pages}{6262--6267}.
\newblock
\urldef\tempurl%
\url{https://doi.org/10.18653/v1/D19-1651}
\showDOI{\tempurl}


\bibitem[Cao et~al\mbox{.}(2018)]%
        {cao2018faithful}
\bibfield{author}{\bibinfo{person}{Ziqiang Cao}, \bibinfo{person}{Furu Wei}, \bibinfo{person}{Wenjie Li}, {and} \bibinfo{person}{Sujian Li}.} \bibinfo{year}{2018}\natexlab{}.
\newblock \showarticletitle{Faithful to the original: Fact aware neural abstractive summarization}. In \bibinfo{booktitle}{\emph{thirty-second AAAI conference on artificial intelligence}}.
\newblock


\bibitem[Cui et~al\mbox{.}(2018)]%
        {cui-etal-2018-neural}
\bibfield{author}{\bibinfo{person}{Lei Cui}, \bibinfo{person}{Furu Wei}, {and} \bibinfo{person}{Ming Zhou}.} \bibinfo{year}{2018}\natexlab{}.
\newblock \showarticletitle{Neural Open Information Extraction}. In \bibinfo{booktitle}{\emph{Proceedings of the 56th Annual Meeting of the Association for Computational Linguistics (Volume 2: Short Papers)}}. \bibinfo{publisher}{Association for Computational Linguistics}, \bibinfo{address}{Melbourne, Australia}, \bibinfo{pages}{407--413}.
\newblock
\urldef\tempurl%
\url{https://doi.org/10.18653/v1/P18-2065}
\showDOI{\tempurl}


\bibitem[Ding et~al\mbox{.}(2021)]%
        {ding2021few}
\bibfield{author}{\bibinfo{person}{Ning Ding}, \bibinfo{person}{Guangwei Xu}, \bibinfo{person}{Yulin Chen}, \bibinfo{person}{Xiaobin Wang}, \bibinfo{person}{Xu Han}, \bibinfo{person}{Pengjun Xie}, \bibinfo{person}{Hai-Tao Zheng}, {and} \bibinfo{person}{Zhiyuan Liu}.} \bibinfo{year}{2021}\natexlab{}.
\newblock \showarticletitle{Few-nerd: A few-shot named entity recognition dataset}.
\newblock \bibinfo{journal}{\emph{arXiv preprint arXiv:2105.07464}} (\bibinfo{year}{2021}).
\newblock


\bibitem[Dong et~al\mbox{.}(2014)]%
        {dong2014knowledge}
\bibfield{author}{\bibinfo{person}{Xin Dong}, \bibinfo{person}{Evgeniy Gabrilovich}, \bibinfo{person}{Geremy Heitz}, \bibinfo{person}{Wilko Horn}, \bibinfo{person}{Ni Lao}, \bibinfo{person}{Kevin Murphy}, \bibinfo{person}{Thomas Strohmann}, \bibinfo{person}{Shaohua Sun}, {and} \bibinfo{person}{Wei Zhang}.} \bibinfo{year}{2014}\natexlab{}.
\newblock \showarticletitle{Knowledge vault: A web-scale approach to probabilistic knowledge fusion}. In \bibinfo{booktitle}{\emph{Proceedings of the 20th ACM SIGKDD international conference on Knowledge discovery and data mining}}. \bibinfo{pages}{601--610}.
\newblock


\bibitem[Fader et~al\mbox{.}(2011)]%
        {fader2011identifying}
\bibfield{author}{\bibinfo{person}{Anthony Fader}, \bibinfo{person}{Stephen Soderland}, {and} \bibinfo{person}{Oren Etzioni}.} \bibinfo{year}{2011}\natexlab{}.
\newblock \showarticletitle{Identifying relations for open information extraction}. In \bibinfo{booktitle}{\emph{Proceedings of the 2011 conference on empirical methods in natural language processing}}. \bibinfo{pages}{1535--1545}.
\newblock


\bibitem[Gashteovski et~al\mbox{.}(2017)]%
        {gashteovski2017minie}
\bibfield{author}{\bibinfo{person}{Kiril Gashteovski}, \bibinfo{person}{Rainer Gemulla}, {and} \bibinfo{person}{Luciano~del Corro}.} \bibinfo{year}{2017}\natexlab{}.
\newblock \showarticletitle{Minie: minimizing facts in open information extraction}. Association for Computational Linguistics.
\newblock


\bibitem[Han et~al\mbox{.}(2023)]%
        {han2023information}
\bibfield{author}{\bibinfo{person}{Ridong Han}, \bibinfo{person}{Tao Peng}, \bibinfo{person}{Chaohao Yang}, \bibinfo{person}{Benyou Wang}, \bibinfo{person}{Lu Liu}, {and} \bibinfo{person}{Xiang Wan}.} \bibinfo{year}{2023}\natexlab{}.
\newblock \showarticletitle{Is Information Extraction Solved by ChatGPT? An Analysis of Performance, Evaluation Criteria, Robustness and Errors}.
\newblock \bibinfo{journal}{\emph{arXiv preprint arXiv:2305.14450}} (\bibinfo{year}{2023}).
\newblock


\bibitem[Kingma and Ba(2014)]%
        {kingma2014adam}
\bibfield{author}{\bibinfo{person}{Diederik~P Kingma} {and} \bibinfo{person}{Jimmy Ba}.} \bibinfo{year}{2014}\natexlab{}.
\newblock \showarticletitle{Adam: A method for stochastic optimization}.
\newblock \bibinfo{journal}{\emph{arXiv preprint arXiv:1412.6980}} (\bibinfo{year}{2014}).
\newblock


\bibitem[Kolluru et~al\mbox{.}(2020a)]%
        {kolluru-etal-2020-openie6}
\bibfield{author}{\bibinfo{person}{Keshav Kolluru}, \bibinfo{person}{Vaibhav Adlakha}, \bibinfo{person}{Samarth Aggarwal}, \bibinfo{person}{{Mausam}}, {and} \bibinfo{person}{Soumen Chakrabarti}.} \bibinfo{year}{2020}\natexlab{a}.
\newblock \showarticletitle{{O}pen{IE}6: {I}terative {G}rid {L}abeling and {C}oordination {A}nalysis for {O}pen {I}nformation {E}xtraction}. In \bibinfo{booktitle}{\emph{Proceedings of the 2020 Conference on Empirical Methods in Natural Language Processing (EMNLP)}}. \bibinfo{publisher}{Association for Computational Linguistics}, \bibinfo{address}{Online}, \bibinfo{pages}{3748--3761}.
\newblock
\urldef\tempurl%
\url{https://doi.org/10.18653/v1/2020.emnlp-main.306}
\showDOI{\tempurl}


\bibitem[Kolluru et~al\mbox{.}(2020b)]%
        {kolluru-etal-2020-imojie}
\bibfield{author}{\bibinfo{person}{Keshav Kolluru}, \bibinfo{person}{Samarth Aggarwal}, \bibinfo{person}{Vipul Rathore}, \bibinfo{person}{{Mausam}}, {and} \bibinfo{person}{Soumen Chakrabarti}.} \bibinfo{year}{2020}\natexlab{b}.
\newblock \showarticletitle{{IM}o{JIE}: Iterative Memory-Based Joint Open Information Extraction}. In \bibinfo{booktitle}{\emph{Proceedings of the 58th Annual Meeting of the Association for Computational Linguistics}}. \bibinfo{publisher}{Association for Computational Linguistics}, \bibinfo{address}{Online}, \bibinfo{pages}{5871--5886}.
\newblock
\urldef\tempurl%
\url{https://doi.org/10.18653/v1/2020.acl-main.521}
\showDOI{\tempurl}


\bibitem[Kolluru et~al\mbox{.}(2022)]%
        {kolluru2022alignment}
\bibfield{author}{\bibinfo{person}{Keshav Kolluru}, \bibinfo{person}{Muqeeth Mohammed}, \bibinfo{person}{Shubham Mittal}, \bibinfo{person}{Soumen Chakrabarti}, {et~al\mbox{.}}} \bibinfo{year}{2022}\natexlab{}.
\newblock \showarticletitle{Alignment-Augmented Consistent Translation for Multilingual Open Information Extraction}. In \bibinfo{booktitle}{\emph{Proceedings of the 60th Annual Meeting of the Association for Computational Linguistics (Volume 1: Long Papers)}}. \bibinfo{pages}{2502--2517}.
\newblock


\bibitem[Papineni et~al\mbox{.}(2002)]%
        {papineni2002bleu}
\bibfield{author}{\bibinfo{person}{Kishore Papineni}, \bibinfo{person}{Salim Roukos}, \bibinfo{person}{Todd Ward}, {and} \bibinfo{person}{Wei-Jing Zhu}.} \bibinfo{year}{2002}\natexlab{}.
\newblock \showarticletitle{Bleu: a method for automatic evaluation of machine translation}. In \bibinfo{booktitle}{\emph{Proceedings of the 40th annual meeting of the Association for Computational Linguistics}}. \bibinfo{pages}{311--318}.
\newblock


\bibitem[Raffel et~al\mbox{.}(2020)]%
        {raffel2020exploring}
\bibfield{author}{\bibinfo{person}{Colin Raffel}, \bibinfo{person}{Noam Shazeer}, \bibinfo{person}{Adam Roberts}, \bibinfo{person}{Katherine Lee}, \bibinfo{person}{Sharan Narang}, \bibinfo{person}{Michael Matena}, \bibinfo{person}{Yanqi Zhou}, \bibinfo{person}{Wei Li}, \bibinfo{person}{Peter~J Liu}, {et~al\mbox{.}}} \bibinfo{year}{2020}\natexlab{}.
\newblock \showarticletitle{Exploring the limits of transfer learning with a unified text-to-text transformer.}
\newblock \bibinfo{journal}{\emph{J. Mach. Learn. Res.}} \bibinfo{volume}{21}, \bibinfo{number}{140} (\bibinfo{year}{2020}), \bibinfo{pages}{1--67}.
\newblock


\bibitem[Roy et~al\mbox{.}(2019)]%
        {roy2019supervising}
\bibfield{author}{\bibinfo{person}{Arpita Roy}, \bibinfo{person}{Youngja Park}, \bibinfo{person}{Taesung Lee}, {and} \bibinfo{person}{Shimei Pan}.} \bibinfo{year}{2019}\natexlab{}.
\newblock \showarticletitle{Supervising unsupervised open information extraction models}. In \bibinfo{booktitle}{\emph{Proceedings of the 2019 Conference on Empirical Methods in Natural Language Processing and the 9th International Joint Conference on Natural Language Processing (EMNLP-IJCNLP)}}. \bibinfo{pages}{728--737}.
\newblock


\bibitem[Saha et~al\mbox{.}(2017)]%
        {saha2017bootstrapping}
\bibfield{author}{\bibinfo{person}{Swarnadeep Saha}, \bibinfo{person}{Harinder Pal}, {et~al\mbox{.}}} \bibinfo{year}{2017}\natexlab{}.
\newblock \showarticletitle{Bootstrapping for numerical open ie}. In \bibinfo{booktitle}{\emph{Proceedings of the 55th Annual Meeting of the Association for Computational Linguistics (Volume 2: Short Papers)}}. \bibinfo{pages}{317--323}.
\newblock


\bibitem[Schmitz et~al\mbox{.}(2012)]%
        {schmitz2012open}
\bibfield{author}{\bibinfo{person}{Michael Schmitz}, \bibinfo{person}{Stephen Soderland}, \bibinfo{person}{Robert Bart}, \bibinfo{person}{Oren Etzioni}, {et~al\mbox{.}}} \bibinfo{year}{2012}\natexlab{}.
\newblock \showarticletitle{Open language learning for information extraction}. In \bibinfo{booktitle}{\emph{Proceedings of the 2012 joint conference on empirical methods in natural language processing and computational natural language learning}}. \bibinfo{pages}{523--534}.
\newblock


\bibitem[Stanovsky et~al\mbox{.}(2018)]%
        {stanovsky-etal-2018-supervised}
\bibfield{author}{\bibinfo{person}{Gabriel Stanovsky}, \bibinfo{person}{Julian Michael}, \bibinfo{person}{Luke Zettlemoyer}, {and} \bibinfo{person}{Ido Dagan}.} \bibinfo{year}{2018}\natexlab{}.
\newblock \showarticletitle{Supervised Open Information Extraction}. In \bibinfo{booktitle}{\emph{Proceedings of the 2018 Conference of the North {A}merican Chapter of the Association for Computational Linguistics: Human Language Technologies, Volume 1 (Long Papers)}}. \bibinfo{publisher}{Association for Computational Linguistics}, \bibinfo{address}{New Orleans, Louisiana}, \bibinfo{pages}{885--895}.
\newblock
\urldef\tempurl%
\url{https://doi.org/10.18653/v1/N18-1081}
\showDOI{\tempurl}


\bibitem[Su(2021)]%
        {zhuiyit5pegasus}
\bibfield{author}{\bibinfo{person}{Jianlin Su}.} \bibinfo{year}{2021}\natexlab{}.
\newblock \bibinfo{booktitle}{\emph{T5 PEGASUS - ZhuiyiAI}}.
\newblock \bibinfo{type}{{T}echnical {R}eport}.
\newblock
\urldef\tempurl%
\url{https://github.com/ZhuiyiTechnology/t5-pegasus}
\showURL{%
\tempurl}


\bibitem[Sun et~al\mbox{.}(2018)]%
        {sun2018logician}
\bibfield{author}{\bibinfo{person}{Mingming Sun}, \bibinfo{person}{Xu Li}, \bibinfo{person}{Xin Wang}, \bibinfo{person}{Miao Fan}, \bibinfo{person}{Yue Feng}, {and} \bibinfo{person}{Ping Li}.} \bibinfo{year}{2018}\natexlab{}.
\newblock \showarticletitle{Logician: a unified end-to-end neural approach for open-domain information extraction}. In \bibinfo{booktitle}{\emph{Proceedings of the Eleventh ACM International Conference on Web Search and Data Mining}}. \bibinfo{pages}{556--564}.
\newblock


\bibitem[Wei et~al\mbox{.}(2022)]%
        {wei2022chain}
\bibfield{author}{\bibinfo{person}{Jason Wei}, \bibinfo{person}{Xuezhi Wang}, \bibinfo{person}{Dale Schuurmans}, \bibinfo{person}{Maarten Bosma}, \bibinfo{person}{Fei Xia}, \bibinfo{person}{Ed Chi}, \bibinfo{person}{Quoc~V Le}, \bibinfo{person}{Denny Zhou}, {et~al\mbox{.}}} \bibinfo{year}{2022}\natexlab{}.
\newblock \showarticletitle{Chain-of-thought prompting elicits reasoning in large language models}.
\newblock \bibinfo{journal}{\emph{Advances in Neural Information Processing Systems}}  \bibinfo{volume}{35} (\bibinfo{year}{2022}), \bibinfo{pages}{24824--24837}.
\newblock


\bibitem[Xue et~al\mbox{.}(2020)]%
        {xue2020mt5}
\bibfield{author}{\bibinfo{person}{Linting Xue}, \bibinfo{person}{Noah Constant}, \bibinfo{person}{Adam Roberts}, \bibinfo{person}{Mihir Kale}, \bibinfo{person}{Rami Al-Rfou}, \bibinfo{person}{Aditya Siddhant}, \bibinfo{person}{Aditya Barua}, {and} \bibinfo{person}{Colin Raffel}.} \bibinfo{year}{2020}\natexlab{}.
\newblock \showarticletitle{mT5: A massively multilingual pre-trained text-to-text transformer}.
\newblock \bibinfo{journal}{\emph{arXiv preprint arXiv:2010.11934}} (\bibinfo{year}{2020}).
\newblock


\bibitem[Yan et~al\mbox{.}(2018)]%
        {yan2018assertion}
\bibfield{author}{\bibinfo{person}{Zhao Yan}, \bibinfo{person}{Duyu Tang}, \bibinfo{person}{Nan Duan}, \bibinfo{person}{Shujie Liu}, \bibinfo{person}{Wendi Wang}, \bibinfo{person}{Daxin Jiang}, \bibinfo{person}{Ming Zhou}, {and} \bibinfo{person}{Zhoujun Li}.} \bibinfo{year}{2018}\natexlab{}.
\newblock \showarticletitle{Assertion-based QA with question-aware open information extraction}. In \bibinfo{booktitle}{\emph{Proceedings of the AAAI Conference on Artificial Intelligence}}, Vol.~\bibinfo{volume}{32}.
\newblock


\bibitem[Yu et~al\mbox{.}(2021)]%
        {yu2021maximal}
\bibfield{author}{\bibinfo{person}{Bowen Yu}, \bibinfo{person}{Yucheng Wang}, \bibinfo{person}{Tingwen Liu}, \bibinfo{person}{Hongsong Zhu}, \bibinfo{person}{Limin Sun}, {and} \bibinfo{person}{Bin Wang}.} \bibinfo{year}{2021}\natexlab{}.
\newblock \showarticletitle{Maximal clique based non-autoregressive open information extraction}. In \bibinfo{booktitle}{\emph{Proceedings of the 2021 Conference on Empirical Methods in Natural Language Processing}}. \bibinfo{pages}{9696--9706}.
\newblock


\bibitem[Zhan and Zhao(2020)]%
        {zhan2020span}
\bibfield{author}{\bibinfo{person}{Junlang Zhan} {and} \bibinfo{person}{Hai Zhao}.} \bibinfo{year}{2020}\natexlab{}.
\newblock \showarticletitle{Span model for open information extraction on accurate corpus}. In \bibinfo{booktitle}{\emph{Proceedings of the AAAI Conference on Artificial Intelligence}}, Vol.~\bibinfo{volume}{34}. \bibinfo{pages}{9523--9530}.
\newblock


\bibitem[Zhou et~al\mbox{.}(2022)]%
        {ijcai2022p793}
\bibfield{author}{\bibinfo{person}{Shaowen Zhou}, \bibinfo{person}{Bowen Yu}, \bibinfo{person}{Aixin Sun}, \bibinfo{person}{Cheng Long}, \bibinfo{person}{Jingyang Li}, {and} \bibinfo{person}{Jian Sun}.} \bibinfo{year}{2022}\natexlab{}.
\newblock \showarticletitle{A Survey on Neural Open Information Extraction: Current Status and Future Directions}. In \bibinfo{booktitle}{\emph{Proceedings of the Thirty-First International Joint Conference on Artificial Intelligence, {IJCAI-22}}}, \bibfield{editor}{\bibinfo{person}{Lud~De Raedt}} (Ed.). \bibinfo{publisher}{International Joint Conferences on Artificial Intelligence Organization}, \bibinfo{pages}{5694--5701}.
\newblock
\urldef\tempurl%
\url{https://doi.org/10.24963/ijcai.2022/793}
\showDOI{\tempurl}
\newblock
\shownote{Survey Track}.


\end{thebibliography}

\end{document}